# Long-distance tiny face detection based on enhanced YOLOv3 for unmanned system

Jia-Yi Chang[1], Yan-Feng Lu[2,*], Ya-Jun Liu[1], Bo Zhou[1], Hong Qiao[2]

[1]School of Information, Beijing City University, Beijing, China
[2]The State Key Laboratory of Management and Control for Complex Systems
Institute of Automation, Chinese Academy of Science, Beijing, China
`yanfeng.lv@ia.ac.cn`

**Abstract.** Remote tiny face detection applied in unmanned system is a challenging work. The detector cannot obtain sufficient context semantic information due to the relatively long distance. The received poor fine-grained features make the face detection less accurate and robust. To solve the problem of long-distance detection of tiny faces, we propose an enhanced network model (YOLOv3-C) based on the YOLOv3 algorithm for unmanned platform. In this model, we bring in multi-scale features from feature pyramid networks and make the features fusion to adjust prediction feature map of the output, which improves the sensitivity of the entire algorithm for tiny target faces. The enhanced model improves the accuracy of tiny face detection in the cases of long-distance and high-density crowds. The experimental evaluation results demonstrated the superior performance of the proposed YOLOv3-C in comparison with other relevant detectors in remote tiny face detection. It is worth mentioning that our proposed method achieves comparable performance with the state of the art YOLOv4[1] in the tiny face detection tasks.

## 1 Introduction

Face detection, as the basis for face tracking, recognition, and synthesis, is a hot topic in computer vision. At present, face detection mainly focuses on solutions with a clear vision, and there is little research on face detection for video sequences with lower resolutions collected by drones and the other unmanned systems. Due to the tiny face area under the drones' down-view scene and the complicated ground scene background during flight, the airborne platform has the problems of poor contrast and occlusion, which makes the accuracy of face detection technology is very low. And it is difficult to guarantee the robustness, which is also one of the most prominent difficulties in the face detection problem of the UAV moving plat-form. For the existing technical difficulties, researchers have carried out a series of studies. At present, the face detection field mainly uses traditional methods or deep learning methods.

With respect to the traditional method, the AdaBoost [2] method was proposed by Viola and Jones in 2001. This method extracts the Haar feature of the face and uses AdaBoost and cascade classifier to train the classifier of the model. It has a fast real-time detection speed in clear positive face images, but is easily affected by environments. Pedro F. Felzenszwalb et al proposed a variable component-based detection algorithm called DPM [3], in 2008. This algorithm has good robustness in face detection. But because it only considers the optimal solution of the local image, the entire detector cannot obtain the optimal result.

With the development of computer vision, especially the application of deep learning has significantly improved the capabilities of image classification and object detection. Deep learning-based face detection algorithms can be roughly divided into two types. The first type is two-stage detectors: Ross Girshick et al applied convolutional neural networks to target detection for the first time and combined with the candidate region proposals. This method named RCNN [4], which enhances the accuracy and operation capability of target detection; on this basis, Girshick then proposed an optimization model Fast R-CNN[5] and Faster R-CNN[6], which both accelerates training and testing speed and increases the mean average precision. But it cannot meet the needs of real-time detection. Another type of target detection algorithm is an implementation of a single-stage detector: Wei Liu et al proposed an SSD[7] algorithm, which encapsulates all calculations in target detection in a network structure, but it has a lower recognition rate in detecting small targets. The YOLO variants of algorithms proposed by Joseph Redmon, et al. YOLOv1[8] and YOLOv2[9] transform the detection task into a regression problem. Nevertheless, the detection effect is not suitable for objects that are close to each other and small target groups. The application of YOLOv3 algorithm has increased the precision in tiny target recognition, but it is only limited to clear and noise-free images. In terms of tiny face detection, its network structure cannot obtain sufficient semantic information of tiny faces, so it has insufficient detection capabilities in low-resolution images. In order to improve the accuracy of YOLOv3[10] for tiny face detection, this paper proposes YOLOv3-C algorithm model.

The YOLOv3-C algorithm uses two max-pooling operations to make small-scale feature maps have better fine-grained features. Besides, we extended the idea of feature pyramids operation, which can get the larger feature maps and obtain richer context semantic features by using the upsampling operation of 8 times downsampling output feature map in YOLOv3 to concatenate with 4 times down-sampling feature map. This improves the sensitivity of the model during the long-distance tiny faces detection. We use the WIDER FACE dataset to build a face database in different situations for offline training. The model is tested using the WIDER FACE test set (containing 3226 small face images). And the results illustrate that this improved algorithm enhances the original YOLOv3's mAP in face detection, and also provides a foundation for the subsequent detection and recognition of long-range human faces.

## 2   Related work

YOLOv3 continues the general idea of YOLOv1 and YOLOv2. The input image is divided into S*S mesh cells. The grid containing the coordinates of the target center in Ground Truth is responsible for predicting the target object. In the prediction, each cell will generate 3 bounding boxes with different initial sizes. The generated output feature maps have three dimensions, two of which are the dimensions of the extracted features, that is, S*S mentioned above; And another dimension represents the depth of the feature map, which is B * (5 + C), where B is the number of bounding boxes predicted by each grid (in YOLOv3, B = 3), C is the number of categories of the bounding box, excluding the background class (in the COCO dataset[11] C = 80), and 5 represents each bounding box has 5 predicted values: $t_x$、$t_y$、$t_w$、$t_h$、$t_o$, including four coordinate information and an objectness score. YOLOv3's mAP can be comparable to RetinaNet, but the speed is about 4 times faster (51ms vs 198ms), which can well meet the requirements of real-time pedestrian detection. The YOLOv3 network structure model is shown in Fig.1.

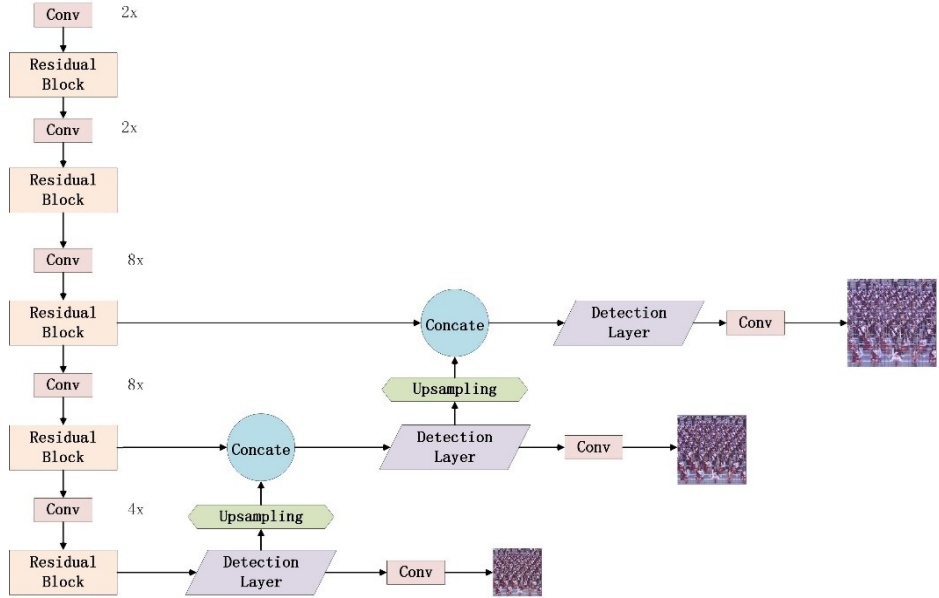

**Fig. 1.** YOLOv3 network model

The Darknet-53 feature extraction network used by YOLOv3 reduces the training difficulty of the network, and it uses 3 different scale feature maps for target prediction, which strengthens the connection between target semantic information and location information. However, the operation of concatenating small-scale feature maps after upsampling with large-scale feature maps cannot make the tiny targets prediction feature maps obtain deeper levels of semantic information. Therefore, the precision decreases during long-distance detection.

## 3    Proposed YOLOv3-C model

Redman uses Darknet-53 as the main network in YOLOv3 to obtain more semantic features in natural scenes and combines the idea of feature pyramids to generate three different scale prediction maps, which greatly improves the accuracy of the network structure for small targets. The YOLOv3 model achieved a detection capability of 57.9AP in 51ms on Titan X. Although YOLOv3 showed better performance, in the actual application of face detection, the phenomenon of missed detection and false detection often occurs due to the problem of dense crowds, long distances and low resolution in the images.

To effectively improve the ability of YOLOv3 to detect faces in dense crowds, the algorithm in this paper focuses on improving the prediction boxes of different scales in YOLOv3. Combining the idea of feature pyramid and feature reorganization, the model can obtain more Fine-grained features of the face from image information.

This paper combines the idea of feature reorganization and feature pyramid. Taking 416 * 416 input images as an example, through the max-pooling operation, to make the feature maps of size 52 * 52 and 26 * 26 have the same size as the feature map of size 13 * 13. And then by concatenation, the predicted feature maps can contain better fine-grained features. At the same time, using the up-sampling operation of 8 times the down-sampling output feature map in YOLOv3 to concatenate with 4 times downsampling feature map. This operation expands one of the output features maps to 104 * 104 size. Besides, a series of 1 * 1 and 3 * 3 convolutions are used to reduce the number of channels and minimize the total amount of calculation to a certain extent. It also improves the non-linear features and expression capabilities of the neural network, which enables the predictive feature maps to obtain richer contextual semantic information, thereby improving sensitivity of long-distance tiny faces. The network structure of the improved YOLOv3-C algorithm is shown in Fig.2.

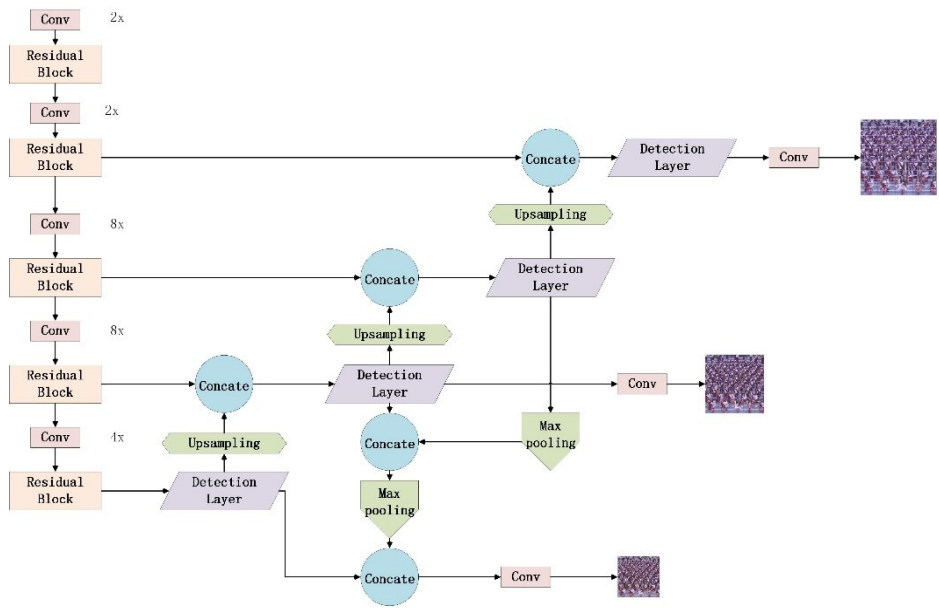

**Fig. 2.** Improved YOLOv3-C network model

The specific improvement details in YOLOv3-C are based on the original output 52 * 52 feature maps, including Up-sampling Operations and Max-pooling Operations:

Up-sampling Operations: The features extracted in the 52 * 52 feature map are concatenation with the features of the neural network layer-11, and used successive 1 * 1 and 3 * 3 convolution operations to obtain a 104 * 104 yolo layer. Through the convolution, we can get a large-scale yolo layer.

Max-pooling Operations: Convolve and pool the feature map once, concatenate it with the neural network layer-87, perform convolution and pooling again, and continue to concatenate with layer-80 to get a 13 * 13 yolo layer. Using this layer to perform a convolution operation to obtain a small-scale yolo layer.

The newly obtained feature layers at two scales are used to perform target classification and position coordinate regression prediction together with the original mesoscale YOLO layer.

## 4 Experiments

### 4.1 Experimental platform configuration

All experiments in the experiments used the GPU on the configuration platform shown in Table 1, including building the required environments such as Cuda10.1, OpenCV3.4, Python, VS2015, as well as running training and testing models. In the experiment, the part about YOLOv3 uses the Darknet-53 framework, and the Faster R-CNN part uses the Tensorflow framework.

Table 1. Experimental platform environ-ment and software and hardware configuration

| Names | Configuration |
| --- | --- |
| CPU | Intel® Core™ I5 |
| GPU | Cuda10.1、Cudnn7.6.5 |
| GPU acceleration rate | NVIDIA GeForce GTX 750 |
| RAM/GB | 8 |
| Operating system | Windows10 |
| Deep learning framework | Darknet-53 / Tensorflow |

### 4.2 Experimental data set preparation and model training

This paper selects the Wider Face data set[12] for model training. The training data set contains a total of 12,877 images. Based on 61 event categories, such as parades, elections, festivals, etc. It includes a total of 159424 human faces. The number of severely blurred faces, which is caused by long-distance, is 95818. And the number of faces with exaggerated expression is 1845, the number of extremely exposed faces is 8134, the number of occlusions in the face part is 2399, and all data sets used in the experiment are completely consistent.

As for model training, YOLOv3 model parameters are used as samples and fine-tune to optimize the training effect of the entire network. The parameters of the initialization model are set to: learning _rate = 0.001, batch = 8, subdivisions = 8, momentum = 0.9, decay = 0.005, max_batches = 10000, policy = steps, scales = 0.1, 0.1.

### 4.3 Evaluation of the YOLOv3-C Method

In this experiment, we compare YOLOv3-C with the original YOLOv3 to show the efficiency and improvement of the improved method.

We used the same test set to evaluate the performance of the YOLOv3 and YOLOv3-C algorithms in the experiment. The test data set contains a total of 3225 images containing tiny faces, and the total cumulative number of ground truth boxes is 39123. Among them, YOLOv3 correctly detected the number of boxes is 22196, and the correct number of boxes detected by YOLOv3-C is 26385. Evaluation indicators[13] include: Recall, average accuracy (mAP), crossover ratio (IoU), and $\frac{RPs}{Img}$. The performance evaluation results are shown in Table 2.

The recall rate is used to evaluate whether the classifier can correctly find all the positive examples and to evaluate whether the prediction is comprehensive. The formula is,

$$\text{Recall} = TP/(TP+FN)*100\% \qquad (1)$$

where, TP indicates the real example, that is, the prediction result is truly the number of faces; FN represents that the prediction result is non-human faces, but the true result is the number of human faces.

The mean average precision (mAP) is the average value of precisions in each category, which is used to evaluate the accuracy of prediction. mAP is calculated as,

$$mAP(\%) = (\Sigma AP)/NC*100\% \qquad (2)$$

where, NC represents the number of all predictable categories in the network. The larger value of it indicates that the algorithm can detect small target faces more accurately.

IoU is the overlap rate of candidate boxes and ground truth, which both generated during the target detection process, that is,

$$IoU_{AB} = (Area\ of\ Overlap)/(Area\ of\ Union) = (A \cap B)/(A+B-A \cap B) \qquad (3)$$

and, if $IoU_{AB}$ is greater than the threshold, the prediction box B is a positive example. When the IoU is larger, the prediction result is better. The threshold for this article is set to the default value of 0.25.

$\frac{RPs}{Img}$ represents the average number of bounding boxes in each image, which can be defined as,

$$RPs/Img = (Total\ detected\ total\ number\ of\ proposals)/(Detected\ image) \qquad (4)$$

Table 2. Performance evaluation of YOLOv3 and YOLOv3-C[1]

| Algorithms | mAP | Recall | IoU | RPs/Img |
| --- | --- | --- | --- | --- |
| YOLOv3 | 44.7 | 56.73 | 49.68 | 88.57 |
| YOLOv3-C | 51.9 | 67.44 | 55.82 | 109.42 |

From the analysis in Table 2, it can be seen that the improved YOLOv3-C has better performance than the original model YOLOv3. The YOLOv3-C algorithm has a 7.2% higher mAP, a 10.71% higher recall rate, and a 6.14% higher intersection ratio than the YOLOv3 algorithm. In YOLOv3-C each image predicted 109.42 bounding boxes, which is much better than in YOLOv3. The results show that the YOLOv3-C algorithm combines the idea of feature reorganization and feature pyramid not only can obtain richer facial features when faced with long-distance face detection, but also has stronger robustness and a good recognition effect in occlusion, exaggerated expressions, ex-treme exposure, blurred vision, etc. Besides, the application of multi-scale feature fusion in the YOLOv3-C algorithm enables the overall model to locate faces more accurate, which greatly improves the algorithm's sensitivity and detection accuracy for tiny targets, and is more suitable for long-distance dense crowd detection. To see the difference between the two models more intuitively, the PR curves of the two are plotted as shown in Fig. 3, and an example Fig.4 and Fig.5 of the test results of YOLOv3 and YOLOv3-C is provided.

---

[1] All evaluation indicators are expressed in percentage system.

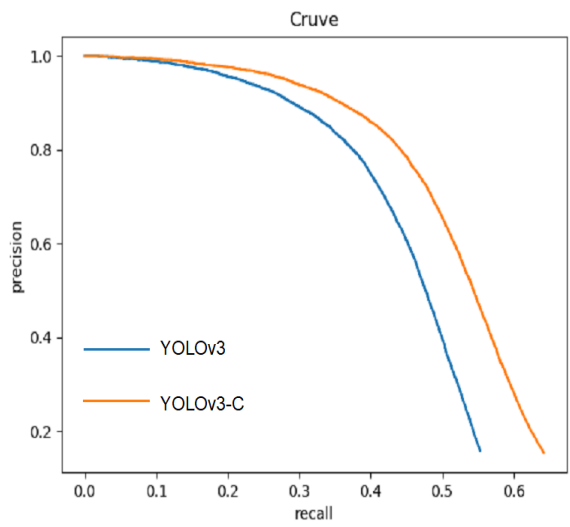

**Fig. 3.** Precision-Recall Rate Curves of YOLOv3 and YOLOv3-C

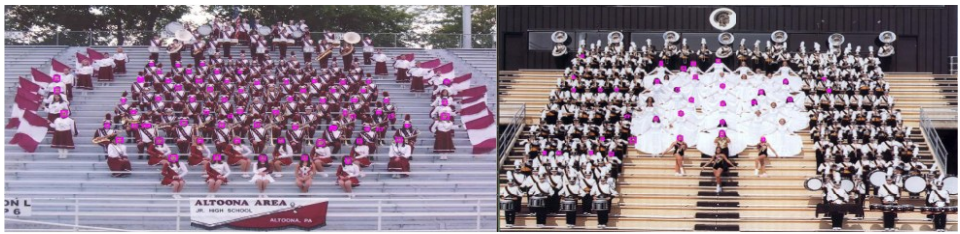

**Fig. 4.** YOLOv3 algorithm

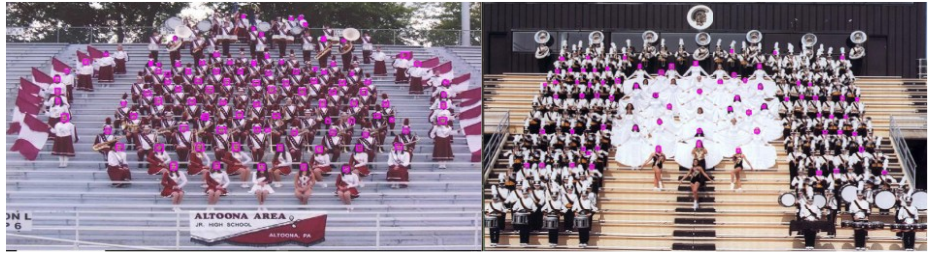

**Fig. 5.** YOLOv3-C algorithm

### 4.4 Comparisons with Other Detection Methods

To evaluate the detection abilities of the proposed method when faced with less training weights, we compare the performance of our proposed method YOLOv3-C with some other methods including the YOLO series of algorithms and the two-stage detector Faster R-CNN, which both trained in the same data set with 4000 iterations.

We adopted these methods because the algorithm in this article is modified based on YOLOv3, which has a good detection effect among single-stage detectors. Faster R-CNN is a representative of the two-stage detector. It used the regional nomination network in extracting prediction frames. It also integrated the operation of proposals regression, classification and extracting features and prediction frames in a network, which achieving an end-to-end target detection model, and gradually becoming the universal target detection model with the highest detection accuracy.

The training results are shown in Table 3. And the P-R curves of various algorithms are plotted in Fig.6.

**Table 3.** Comparison result with other detection methods

| Iteration | Methods | mAP |
|---|---|---|
| 4000 | Faster R-CNN | 31.62 |
| | YOLOv1 | 0.015 |
| | YOLOv2 | 0.016 |
| | YOLOv3 | 42.2 |
| | YOLOv4 | 43.5 |
| | YOLOv3-C | **43.8** |

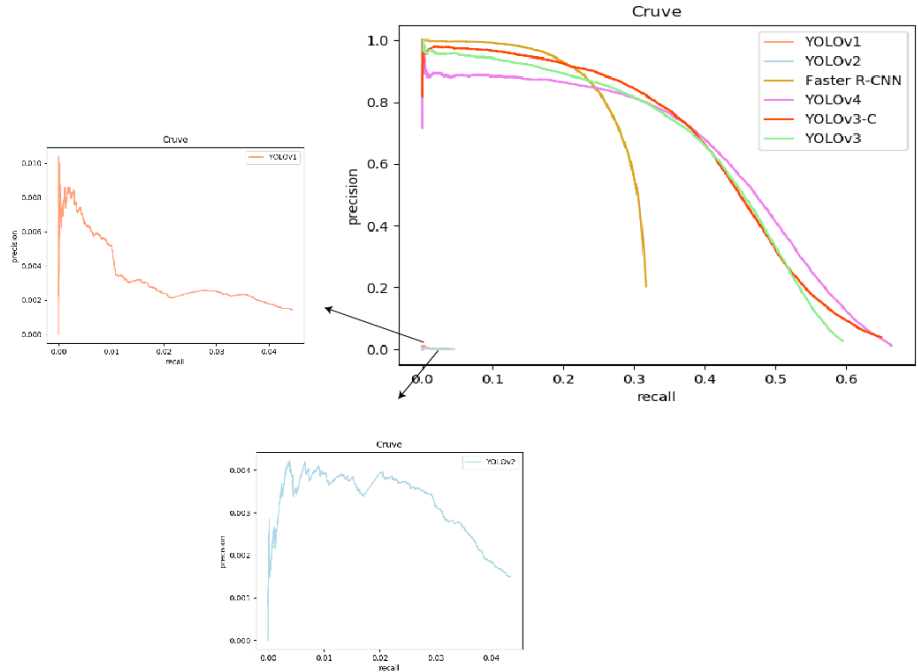

**Fig. 6.** P-R curves of various long-distance face detection models

As shown in Table 3 and Figure 5: YOLOv1, and YOLOv2 have poor performance in long-distance face detection and the results are much lower than the improved method. After training 4000 times, the experimental results verified the superiority of YOLOv3-M compared to YOLOv4. During the training of YOLOv3-M, Faster R-CNN algorithms, the accuracy rate increases with the number of iterations, and the detection capability of the YOLOv3-C algorithm is gradually better than Faster R-CNN. And it has an 12.18% higher accuracy rate than Faster R-CNN. Because YOLOv1 and YOLOv2 have insufficient expressiveness in face detection at long distances, it is impossible to detect tiny faces in lower resolution images. And Therefore, the test results in the comparison experiment only include YOLOv3-C and Faster R-CNN, as shown in Fig.7 and Fig.8. The results show that the improved algorithm proposed in this paper has greatly improved the sensitivity to tiny target faces. When faced with variable environments such as blur, and expression and face exposure, YOLOv3-C has a better comprehensive performance in long-distance and high-density face detection.

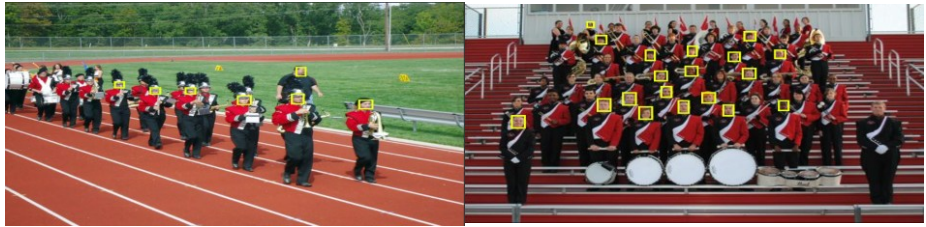

**Fig. 7.** Faster R-CNN algorithm

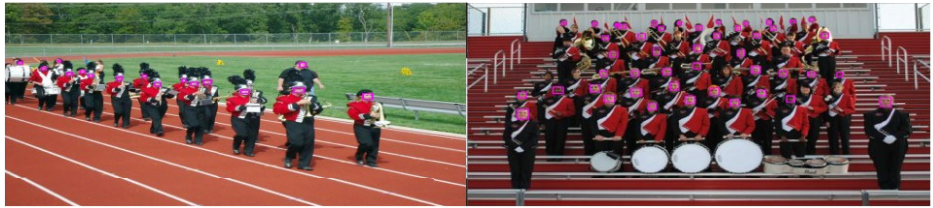

**Fig. 8.** YOLOv3-C algorithm

## 5 Conclusion

Aiming at long-distance tiny faces detection, we proposed an improved algorithm YOLOv3-C based on the end-to-end and real-time object detection YOLOv3 algorithm. The improved method adjusted the structure of the neural network and combined with the ideas of feature map fusion and feature reorganization. The experiments in this paper fully show that the improved algorithm has better sensitivity and precision for long-distance tiny faces detection, which can be used in unmanned system, i.e., UAV.

Although the improved algorithm has been improved to a certain extent, average precision still has not reached a high standard, and still exits a certain number of missed detections. Therefore, in order to achieve better robustness in tiny face detection, the future research will focus on cleaning the tiny face data set, adjusting the cluster anchor boxes generated by K-means, and optimizing the loss function in the algorithm.

## 6 Acknowledgment

This work was supported by the National Natural Science Foundation of China(Grants 61603389, 91648205), and partially supported by the Science and Technology General